\pdfoutput=1
\documentclass{article}

\usepackage{hyperref}

\usepackage{placeins}

\usepackage[preprint]{acl}
\usepackage{times}
\usepackage{latexsym}

\usepackage[T1]{fontenc}
\usepackage{microtype}

\usepackage{inconsolata}

\usepackage{amsfonts}
\usepackage{nicefrac}       
\usepackage{amsmath}
\usepackage{amsthm}
\usepackage{amssymb}
\usepackage{cancel}
\usepackage{mathtools}
\usepackage{bbm}
\usepackage{bm}

\usepackage{cases}         

\usepackage{mathalpha}

\usepackage{microtype}      
\usepackage{enumitem}

\usepackage{pifont}

\usepackage{stmaryrd}

\newcommand{\bb}[1]{\mathbbl{#1}}

\definecolor{lightgray}{RGB}{230, 230, 230}
\definecolor{mathred}{RGB}{204, 69, 90}
\definecolor{mathblue}{RGB}{4, 78, 112}
\definecolor{mathgreen}{RGB}{1, 135, 70}

\definecolor{pennblue}      {cmyk}{1,0.65,0,0.30}
\definecolor{pennred}       {cmyk}{0,1,0.65,0.34}
\definecolor{mygreen}       {rgb} {0.10,0.50,0.10}
\definecolor{mylightgreen}  {rgb} {0.80,0.90,0.80}
\definecolor{lightred}      {rgb} {1.00,0.80,0.80}
\definecolor{lightblue}     {rgb} {0.80,0.80,1.00}
\definecolor{mediumred}     {rgb} {1.00,0.60,0.60}
\definecolor{mediumblue}    {rgb} {0.60,0.60,1.00}

\newcommand{\red}[1]         {{\color{red}#1}}

\newcommand{\bulletcolor}[1] {{\color{pennblue}#1}}

\def\bbX{{\ensuremath{\mathbf X}}}

\def\bbl{{\ensuremath{\mathbf l}}}

\def\bbp{{\ensuremath{\mathbf p}}}

\def\bb0{{\ensuremath{\mathbf 0}}}
\def\bb1{{\ensuremath{\mathbf 1}}}

\def\Diag{{\ensuremath{\text{Diag}}}}

\usepackage{hyperref}
\usepackage{url}
\usepackage{tikz}
\usetikzlibrary{3d}
\usepackage{multirow}
\def \pathancillaries {anciliaries}
\usepackage{amsthm}
\usepackage{epsfig} 
\usepackage{amsfonts}
\usepackage{amsbsy} 
\usepackage{amsmath}
\usepackage{tcolorbox}
\usepackage{mathtools}
\usepackage{lastpage}
\usepackage{multimedia}
\usepackage{calc}
\usepackage{ulem}

\usepackage{subcaption}
\usepackage{pgf} 
\usepackage{tikz}
\usepackage{pgfplots}
\usepackage{transparent}

\usetikzlibrary{shapes} 
\usetikzlibrary{arrows} 
\usetikzlibrary{automata}

\definecolor{BGcolor}{RGB}{240,240,240}
\usepackage{relsize}
\usepackage{listings}
\input{\pathancillaries/penn_colors.sty}
\input{\pathancillaries/my_symbol.sty}
\input{\pathancillaries/my_beamer_defs.sty}
\input{\pathancillaries/my_tikz_defs.sty}

\newcommand \lighturl[1]{{\color[rgb]{0.60,0.60,1.0}\url{#1}}}

\pgfmathsetmacro{\width}{5} 
\pgfmathsetmacro{\nrows}{10} 
\pgfmathsetmacro{\ncolumns}{20} 
\usepackage{dblfloatfix} 

\def\myfactor{0.85}  \def\unit{\myfactor cm}

\def\colors{{"023FA5", "7D87B9", "BEC1D4", "D6BCC0", "BB7784", 
             "8E063B", "4A6FE3", "8595E1", "B5BBE3", "E6AFB9",
             "E07B91", "D33F6A", "11C638", "8DD593", "C6DEC7", 
             "EAD3C6", "F0B98D", "EF9708", "0FCFC0", "9CDED6",
             "D5EAE7", "F3ELEB", "F6C4E1", "F79CD4"}}

\usepackage[utf8]{inputenc} 
\usepackage[T1]{fontenc}    
\usepackage{hyperref}       
\usepackage{url}            
\usepackage{booktabs}       
\usepackage{amsfonts}       
\usepackage{nicefrac}       
\usepackage{microtype}      
\usepackage{xcolor}         
\usepackage{subcaption}

\definecolor{mycyan}{RGB}{0,204,204}

\title{LoRTA: Efficient Low Rank Tensor Adaptation of Large Language Models}

\author{%
  Ignacio Hounie \\
  Univ. of Pennsylvania
  \And
  Charilaos Kanatsoulis\\
  Stanford University
  \And
  Arnuv Tandon\\
  Stanford University
  \And
  Alejandro Ribeiro \\
  Univ. of Pennsylvania
}

\begin{document}

\maketitle
\begin{abstract}
Low Rank Adaptation (LoRA) is a popular Parameter Efficient Fine Tuning (PEFT) method that effectively adapts large pre-trained models for downstream tasks. LoRA parameterizes model updates using low-rank matrices at each layer, significantly reducing the number of trainable parameters and, consequently, resource requirements during fine-tuning. However, the lower bound on the number of trainable parameters remains high due to the use of the low-rank matrix model. Recent works have addressed this limitation by proposing low rank tensor parameterizations for model updates. However, they only exploit redundancy across layers, or tensorize individual matrices using ad-hoc schemes that introduce additional hyperparameters. In this work, we propose a higher-order Candecomp/Parafac (CP) decomposition, enabling a more compact and flexible representation compared to existing matrix and tensor based PEFT methods. The proposed low rank tensor model can reduce the number of trainable parameters, while also allowing for finer-grained control over adapter size. Our experiments on Natural Language Understanding, Instruction Tuning, Preference Optimization and Protein Folding benchmarks demonstrate that our method can achieve a reduction in the number of parameters while maintaining comparable performance.
\end{abstract}

\section{Introduction}

The advent of Large Language Models (LLMs) -- billion parameter scale models pre-trained on vast corpora of data -- has enabled unprecedented capabilities across a wide range of tasks. However, as LLM sizes continue to grow exponentially, their computational and memory demands represent significant challenges, particularly for those lacking access to high-performance computing infrastructure~\citep{varoquaux2024hypesustainabilitypricebiggerisbetter}. This has spurred interest in parameter efficient fine tuning (PEFT) techniques~\citep{2023peftsurvey}, which facilitate the adaptation of LLMs to specific applications, downstream tasks or user preferences, by using only a small fraction of trainable parameters. Most importantly, they reduce GPU memory requirements, primarily by shrinking optimizer states~\citep{Liao2023MemoryEfficient}. Moreover, they provide greater efficiency  in storage and deployment, enabling the management of multiple fine-tuned LLMs with reduced storage footprints and faster load times~\citep{2023thousandsofconcurrent, wen2024batched}, which is particularly relevant for applications requiring rapid model switching across numerous task- or user-specific models.

Beyond computational benefits, PEFT techniques can also mitigate  overfitting risks associated with fine-tuning high-capacity LLMs. By constraining model updates, PEFT methods can act as an implicit regularization mechanism, improving generalization~\citep{2023PEFTeffectiveness, sun2023exploring}. Parameter sharing, a well-established technique in deep learning architecture design, has been shown to improve generalization  across various tasks such as protein folding~\citep{Jumper2021AlphaFold,Lin2021ESMFold}, image segmentation~\citep{ronneberger2015unet}, and generative modeling~\citep{rombach2022high}. Incorporating parameter sharing in PEFT methods has also improved performance in specialized applications with limited data, such as in medical domains~\citep{ dutt2023peft-medical, zhu2024medical-melo}.

Low Rank Adaptation (LoRA) is a popular PEFT approach that uses a low rank parameterization of weight matrix updates~\citep{hu2021lora}. For instance, these allow to fine tune a 175 billion parameter LLM using only 5 million trainable parameters~\citep{hu2021lora} without performance degradation. Since the model updates can be merged with the frozen weights, LoRA incurs no additional inference cost when deployed, unlike prompt~\citep{2021prompt, 2023prompt} and adapter-based~\citep{houlsby2019parameter, 2021adapter, pfeiffer2020adapterfusion} PEFT methods.

However, the lower bound on trainable parameters often remains substantial for large-scale models. 
Recent works have aimed to further reduce the number of parameters in LoRA by using shared pseudorandom low rank projections~\citep{ zhang2023lorafa, 2023vera}, or parameterizing low rank matrices using a pseudorandom basis~\citep{koohpayegani-nola-2024}. We show that the parameter-sharing schemes in these methods can be interpreted as low-rank tensor models with fixed random factors. 

On the other hand, ~\cite{yang2024loretta} leverage low rank tensor adapters by treating each weight update as a tensor with arbitrary dimensions. However, this tensorization scheme not only introduces additional hyperparameters but also forfeits structural information and potential correlations among different weights. Other low-rank tensor adapter models recently proposed for vision transformers~\citep{2023-tensor-adapters-cnn-aaai, 2023-tensor-adapters-cnn-robotics} and LLMs~\citep{ bershatsky2024lotr} treat model layers as an explicit mode, but do not exploit redundancies across attention matrices or heads. Moreover, they use tucker and tensor train models which are less parameter efficient and parsimonious than CANDECOMP/PARAFAC (CP) models~\cite{kolda2009tensor}. 

Building on these insights, we propose LoRTA, a 5th-order CP-based low-rank factorization that unifies parameter updates across layers, heads, and attention matrices. To the best of our knowledge this is the first tensor based method to (i) exploit redundancy in weight updates across layers, heads, and attention matrices by representing updates as a unified 5th-order low-rank tensor (ii) employ a CP model.

We evaluate our method on diverse benchmarks including Natural Language Understanding, Instruction Tuning, Preference Optimization, and Protein Folding. Our experiments demonstrate that LoRTA can achieve up to an order of magnitude reduction in the number of trainable parameters compared to state-of-the-art PEFT methods, with minimal performance trade-offs.

\section{Preliminaries}\label{sec:preliminaries}
\subsection{Transformer Architecture}
We focus on the transformer architecture, although it can be naturally extended to other architectures such as Convolutional Neural Networks and Long Short Term Memory networks. We adopt the problem setting presented in~\citep{thickstun2021transformer-eqs}.  In the transformer model, an initial embedding layer maps input tokens to $d-$dimensional vector representations. These embeddings then pass through a series of layers, each performing multi-head attention, normalization and feed-forward operations. The input to the $l-$th layer of the transformer is a matrix $\bm X^{(l)}\in\mathbb{R}^{N\times d}$, where $N$ is the number of queries, represented in a $d-$dimensional feature space. A vanilla transformer layer with $H$ attention heads is then defined as follows:

\begin{align*}
 \bm X^{(l+1)} & =\texttt{LayerNorm}\left(\bm{Y}^{(l)}+\texttt{MLP}\left(\bm{Y}^{(l)}\right)\right)\\
\bm{Y}^{(l)} & =\texttt{LayerNorm}\left(\bm{X}^{(l)}+\texttt{Attn}\left(\bm{X}^{(l)}\right)\right)\\
    \texttt{Attn}\left(\bm X^{(l)}\right)& = \bm X^{(l)} \\ &\hspace{-20mm}+\sum_{h=1}^H\texttt{softmax}\left(\frac{\bm X^{(l)}\bm Q_h^{(l)}\bm K_h^{(l)^T}\bm X^{(l)^T}}{\sqrt{d}}\right)\bm X^{(l)}\bm V_h^{(l)} \bm P_h^{(l)^T}\\
    \texttt{MLP}\left(\bm X^{(l)}\right)&=\texttt{ReLU}\left(\bm X^{(l)} \bm G_1^T + \bm 1_N\bm b_1^T\right)\bm G_2^T+ \bm 1_N\bm b_2^T,
\end{align*}

where $\bm K_h^{(l)},\bm Q_h^{(l)},\bm V_h^{(l)},\bm P_h^{(l)}\in\mathbb{R}^{d\times d_H}$ are the key, query, value and projection matrices respectively, for head $h$ and layer $l$. 

\subsection{Low Rank (matrix) Adaptation}


LoRA modifies the pre-trained weights by adding a trainable update. Explicitly, at each layer and head $h$:
\begin{align}
    \bm K_h =\bm K_h^0 + d\bm {K}_h,\quad 
    \bm Q_h =\bm Q_h^0 +d\bm {Q}_h,\nonumber\\
    \bm V_h =\bm V_h^0 +d \bm {V}_h,\quad
    \bm P_h=\bm P_h^0 +d\bm {P}_h,\nonumber
\end{align}
where $\bm K^0,\bm Q^0,\bm V^0,\bm P^0$ denote the pre-trained weights and $ d\bm K, d\bm Q,d \bm V,d\bm P$ the trainable adapters.

While each attention head's MLP contains two trainable matrices, $ \bm G_1$ and $ \bm G_2$, our focus is on fine-tuning the attention matrices. This has been demonstrated to be effective for LLM adaptation~\citep{hu2021lora, 2023vera}. Nevertheless, these methods can be easily extended to other parameters, including the MLP weights.

Let  $\bm W_h\in\left\{{\bm Q_h}, {\bm K_h}, {\bm V_h}, {\bm P_h}\right\}$ for $h=1,\dots,H$ denote the query, key, value and projection matrices, respectively, for each attention head. After  concatenating updates across all attention heads, we get: 
\begin{align*}
	d\tilde{\bm W} = \left(d\bm W_1, \ldots, d\bm  W_H\right)\in \mathbb{R}^{d\times d}.
\end{align*}
\cite{hu2021lora} proposed to parametrize the updates using rank-$r$ matrices, which can be expressed as
\begin{align}\label{eq:lora}
 d\tilde{\bm W} = \frac{\alpha}{r}\bm A \bm B^T, \quad \bm A, \bm B \in \mathbb{R}^{d\times r},
\end{align}

 where $\alpha$ is a constant and $r$ denotes the rank of the update. The scaling factor simply aims to reduce the efforts of re-tuning the learning rate when training adapters of varying rank. It has been shown that while this scaling heuristic works well for smaller ranks, it can be sub-optimal for larger ranks~\citep{kalajdzievski2023rs-lora}.~\cite{hayou2024lora+} have also shown that setting the learning rate for the $\bm A$ and $\bm B$ matrices appropriately can further improve convergence and performance.

Although LoRA is an efficient fine-tuning technique, the number of parameters required for each layer is at least $8\cdot d\cdot r$.  Thus, the total number of trainable parameters is:
\begin{equation}
    \# \text{parameters}\left(\text{LoRA}\right) = 2\cdot M \cdot d \cdot L\cdot r,
\end{equation}
where $L$ is the total number of layers and $M$ the number of finetuned attention/projection matrices. Even with $r=1$, this results in $4\cdot M \cdot d\cdot L$ parameters. In practice, for 
LLMs with high dimensionality ($d$) and many layers ($L$), this lower bound can still lead to a significant number of trainable parameters.

LoRA has also been combined with model weight quantization~\citep{dettmers2024qlora}, further decreasing resource requirements. 
Unlike adapter-based PEFT methods~\citep{houlsby2019parameter, pfeiffer2020adapterfusion, ruckle2020adapterdrop, 2021adapter}, LoRA does not introduce additional inference time overhead during deployment, as the trainable matrices can be integrated with the fixed weights.

Building upon this foundation, AdaLoRA~\citep{zhang2023adaptive} expands the LoRA technique by introducing dynamic rank adjustment for low-rank matrices during fine-tuning. The fundamental concept involves optimally allocating the parameter resources by selectively pruning less crucial components of the matrices based on an importance metric. LoRA-FA~\citep{zhang2023lorafa} reduces the number of trainable parameters by freezing the $\bm A$ matrix to its random initialisation, while achieving similar performance to LoRA.

\subsection{Tensor Algebra}
In the following sections we introduce our proposed LoRTA framework, which is a tensor adaptation model for PEFT. To facilitate the upcoming analysis, we briefly present some tensor algebra preliminaries and refer the reader to Appendix \ref{App:tensors} and \cite{sidiropoulos2017tensor,kolda2009tensor} for further details.
	
A $N$-order tensor $\mathcal{\bm X}\in\mathbb{R}^{I_1\times I_2\times\dots\times I_N}$ is an $N$-way array indexed by $i_1,i_2,\dots,i_N$ with elements $\mathcal{\bm X}(i_1,i_2,\dots,i_N)$. 
It consists of $N$ types of modes: $\mathcal{\bm X}(:,i_2,\dots,i_N)$, $\mathcal{\bm X}(i_1,:,\dots,i_N),\dots,\mathcal{\bm X}(i_1,i_2,\dots,:)$. Any tensor can be decomposed as a sum of $N$-way outer products as: 
	 where $\bm{A_n}=[\bm a_n^1,\bm a_n^2,\dots,\bm a_n^R]\in\mathbb{R}^{I_n\times R},~n=1,\dots,N$ are called the low rank factors of the tensor. The above expression represents the \textit{canonical polyadic decomposition} (CPD) or \textit{parallel factor analysis} (PARAFAC)~\citep{harshman1994parafac} of a tensor. A tensor can be fully characterized by its latent factors, so we can represent a tensor by its CPD model as:
  \begin{equation}
      \mathcal{\bm X} =\left\llbracket{\bm A_1},{\bm A_2},\dots,{\bm A_N}\right\rrbracket.
  \end{equation}

 Unlike other tensor models, such as Tucker and Block Term Decomposition (BTD), the CPD model is unique under certain conditions. As a result, the CPD model is often preferred when the goal is to minimize the number of parameters.

A tensor can also be represented as a set of matrices, by fixing all the modes but two as:
  \begin{align}
      &\mathcal{\bm X}\left[:,:,i_3,\dots,i_N\right] = \\ & \; {\bm A_1}\left(\Diag\left(\bm A_3\left(i_3,:\right)\right)\odot\cdots\odot\Diag\left(\bm A_N\left(i_N,:\right)\right)\right){\bm A_2^T},
  \end{align}  
where $\Diag\left(\bm A_n\left(i_n,:\right)\right)$ is the diagonal matrix with diagonal equal to $\bm A_N\left(i_n,:\right)$.
 \begin{figure*}[hbt!]
    \centering
    \begin{subfigure}[b]{0.4\textwidth}
        \centering
        \usetikzlibrary{3d}

\pgfmathsetmacro{\width}{5}      
\pgfmathsetmacro{\featuresep}{0} 
\pgfmathsetmacro{\nx}{8} 
\pgfmathsetmacro{\ny}{3} 
\pgfmathsetmacro{\nz}{5} 

\begin{tikzpicture}[z  = {(0.5cm,0.5cm)},
                    x  = {(0.95cm,-0.25cm)},
                    y  = {(0cm,0.9cm)},
                    scale = 0.08]

\pgfmathsetmacro{\lx}{int(\nx-1)} 
\pgfmathsetmacro{\ly}{int(\ny-1)} 
\pgfmathsetmacro{\lz}{int(\nz-1)} 

\begin{scope}[opacity = 0.8]

\foreach \z in {0, ..., \lz}
{
   \pgfmathsetmacro{\thecurrentcolor}{\colors[mod(\z*5,40)]}
   \definecolor{currentcolor}{HTML}{\thecurrentcolor}

   \foreach \y in {0, ..., \ly}
   {
      \foreach \x in {0, ..., \lx}
      {

         \pgfmathsetmacro{\colorstrength}{random(100)}         
         
         \pgfmathsetmacro{\posx}{ (\width + 0          )*\x}
         \pgfmathsetmacro{\posy}{ (\width + 0          )*\y}
         \pgfmathsetmacro{\posz}{-(\width + \featuresep)*\z}

         \path [draw, fill = currentcolor!\colorstrength] 
             ( \posx,   \posy,   \posz )    ++
             ( 0,       0,       \width) -- ++           
             ( \width,  0,       0     ) -- ++ 
             ( 0,       \width,  0     ) --++ 
             (-\width,  0,       0     ) --++ 
             (0,       -\width,  0     );
         \path [draw, fill = currentcolor!\colorstrength] 
             ( \posx,   \posy,   \posz )    ++
             ( 0,       0,       0     ) -- ++           
             ( 0,       0,       \width) -- ++ 
             ( 0,       \width,  0     ) --++ 
             ( 0,       0,      -\width) --++ 
             (0,       -\width,  0     );
         \path [draw, fill = currentcolor!\colorstrength] 
             ( \posx,   \posy,   \posz )    ++
             ( 0,       0,       0     ) -- ++           
             ( 0,       0,       \width) -- ++ 
             ( \width,  0,       0     ) --++ 
             ( 0,       0,      -\width) --++ 
             (-\width,  0,       0     );
         \path [draw, fill = currentcolor!\colorstrength] 
             ( \posx,   \posy,   \posz )    ++
             ( 0,       0,       0     ) -- ++           
             ( \width,  0,       0     ) -- ++ 
             ( 0,       \width,  0     ) --++ 
             (-\width,  0,       0     ) --++ 
             (0,       -\width,  0     );
         \path [draw, fill = currentcolor!\colorstrength] 
             ( \posx,   \posy,   \posz )    ++
             ( \width,  0,       0     ) -- ++           
             ( 0,       0,       \width) -- ++ 
             ( 0,       \width,  0     ) --++ 
             ( 0,       0,      -\width) --++ 
             (0,       -\width,  0     );
         \path [draw, fill = currentcolor!\colorstrength] 
             ( \posx,   \posy,   \posz )    ++
             ( 0,       \width,  0     ) -- ++           
             ( 0,       0,       \width) -- ++ 
             ( \width,  0,       0     ) --++ 
             ( 0,       0,      -\width) --++ 
             (-\width,  0,       0     );
      }
   }

   \foreach \z in {0, ..., \lz}
    {
   \pgfmathsetmacro{\thecurrentcolor}{\colors[23]}
   \definecolor{currentcolor}{HTML}{\thecurrentcolor}
   \foreach \y in {0, ..., \ly}
   {
         \pgfmathsetmacro{\colorstrength}{20+random(80)}         
         
         \pgfmathsetmacro{\posx}{ (\width + 0          )*\nx + 1*\width }
         \pgfmathsetmacro{\posy}{ (\width + 0          )*\y}
         \pgfmathsetmacro{\posz}{-\width *\z}

         \path [draw, fill = currentcolor!\colorstrength] 
             ( \posx,   \posy,   \posz )    ++
             ( 0,       0,       \width) -- ++           
             ( \width,  0,       0     ) -- ++ 
             ( 0,       \width,  0     ) --++ 
             (-\width,  0,       0     ) --++ 
             (0,       -\width,  0     );
         \path [draw, fill = currentcolor!\colorstrength] 
             ( \posx,   \posy,   \posz )    ++
             ( 0,       0,       0     ) -- ++           
             ( 0,       0,       \width) -- ++ 
             ( 0,       \width,  0     ) --++ 
             ( 0,       0,      -\width) --++ 
             (0,       -\width,  0     );
         \path [draw, fill = currentcolor!\colorstrength] 
             ( \posx,   \posy,   \posz )    ++
             ( 0,       0,       0     ) -- ++           
             ( 0,       0,       \width) -- ++ 
             ( \width,  0,       0     ) --++ 
             ( 0,       0,      -\width) --++ 
             (-\width,  0,       0     );
         \path [draw, fill = currentcolor!\colorstrength] 
             ( \posx,   \posy,   \posz )    ++
             ( 0,       0,       0     ) -- ++           
             ( \width,  0,       0     ) -- ++ 
             ( 0,       \width,  0     ) --++ 
             (-\width,  0,       0     ) --++ 
             (0,       -\width,  0     );
         \path [draw, fill = currentcolor!\colorstrength] 
             ( \posx,   \posy,   \posz )    ++
             ( \width,  0,       0     ) -- ++           
             ( 0,       0,       \width) -- ++ 
             ( 0,       \width,  0     ) --++ 
             ( 0,       0,      -\width) --++ 
             (0,       -\width,  0     );
         \path [draw, fill = currentcolor!\colorstrength] 
             ( \posx,   \posy,   \posz )    ++
             ( 0,       \width,  0     ) -- ++           
             ( 0,       0,       \width) -- ++ 
             ( \width,  0,       0     ) --++ 
             ( 0,       0,      -\width) --++ 
             (-\width,  0,       0     );
   }
   }

}
    \pgfmathsetmacro{\thecurrentcolor}{\colors[23]}
   \definecolor{currentcolor}{HTML}{\thecurrentcolor}
   \foreach \x in {0, ..., \lx}
   {
         \pgfmathsetmacro{\colorstrength}{20+random(80)}         
         
         \pgfmathsetmacro{\posx}{ (\width + 0          )*\x}
         \pgfmathsetmacro{\posy}{ (\width + 0          )*(\ny+1)}
         \pgfmathsetmacro{\posz}{-(\width + \featuresep)*\lz}

         \path [draw, fill = currentcolor!\colorstrength] 
             ( \posx,   \posy,   \posz )    ++
             ( 0,       0,       \width) -- ++           
             ( \width,  0,       0     ) -- ++ 
             ( 0,       \width,  0     ) --++ 
             (-\width,  0,       0     ) --++ 
             (0,       -\width,  0     );
         \path [draw, fill = currentcolor!\colorstrength] 
             ( \posx,   \posy,   \posz )    ++
             ( 0,       0,       0     ) -- ++           
             ( 0,       0,       \width) -- ++ 
             ( 0,       \width,  0     ) --++ 
             ( 0,       0,      -\width) --++ 
             (0,       -\width,  0     );
         \path [draw, fill = currentcolor!\colorstrength] 
             ( \posx,   \posy,   \posz )    ++
             ( 0,       0,       0     ) -- ++           
             ( 0,       0,       \width) -- ++ 
             ( \width,  0,       0     ) --++ 
             ( 0,       0,      -\width) --++ 
             (-\width,  0,       0     );
         \path [draw, fill = currentcolor!\colorstrength] 
             ( \posx,   \posy,   \posz )    ++
             ( 0,       0,       0     ) -- ++           
             ( \width,  0,       0     ) -- ++ 
             ( 0,       \width,  0     ) --++ 
             (-\width,  0,       0     ) --++ 
             (0,       -\width,  0     );
         \path [draw, fill = currentcolor!\colorstrength] 
             ( \posx,   \posy,   \posz )    ++
             ( \width,  0,       0     ) -- ++           
             ( 0,       0,       \width) -- ++ 
             ( 0,       \width,  0     ) --++ 
             ( 0,       0,      -\width) --++ 
             (0,       -\width,  0     );
         \path [draw, fill = currentcolor!\colorstrength] 
             ( \posx,   \posy,   \posz )    ++
             ( 0,       \width,  0     ) -- ++           
             ( 0,       0,       \width) -- ++ 
             ( \width,  0,       0     ) --++ 
             ( 0,       0,      -\width) --++ 
             (-\width,  0,       0     );
   }

{
    {
      \pgfmathsetmacro{\poszlabel}{-(\width + \featuresep)*\lz}
      \node at (-1*\width, \ly*\width+4*\width, \poszlabel-0.5*\width) [anchor=north west] {$\mathbf{a}$};
   }
   {
   \pgfmathsetmacro{\posxlabel}{ (\width + 0 )*\nx + 2*\width }
      \foreach \z in {0, ..., \lz} {
          \pgfmathsetmacro{\poszlabel}{-(\width + \featuresep)*\z}
          \node at (\posxlabel, -\width, \poszlabel) [anchor=south west] {$\mathbf{b}_{\z}$};
       }
   }
   }
   {
   \pgfmathsetmacro{\zaxispos}{-(\lz*(\width+\featuresep)}
   \path [draw, -stealth, thick, black, above] 
            (-2*\width,0,\zaxispos-2*\width) -- ++ (4*\width , 0, 0) node {$d$};

\path [draw, -stealth, thick, black, above] 
            (-2*\width,0,\zaxispos-2*\width) -- ++ (0,\ny*\width, 0) node {$d_H$};

\path [draw, -stealth, thick, black, above] 
            (-2*\width,0,\zaxispos-2*\width) -- ++ (0, 0, -\zaxispos-\width) node {Heads};
    }
\end{scope}

\end{tikzpicture}
        \caption{LoRA.}
    \end{subfigure}
    \hspace{0.05\textwidth}
    \begin{subfigure}[b]{0.4\textwidth}
        \centering
        \input{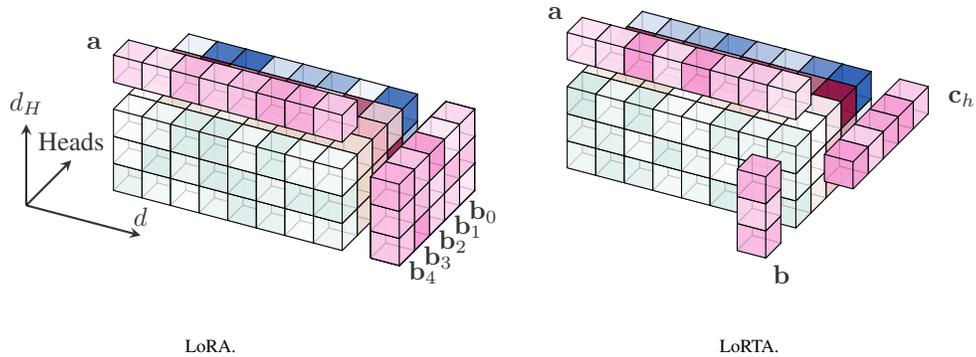}
    \caption{LoRTA.}
    \end{subfigure}
    \caption{Illustration of a rank 1 adapter for a single weight matrix with multiple heads. (Left) The LoRA update for head $h$ is computed as $d \bm W_h =  \bm b_h \circ \bm a$. (Right) The update using a third order low rank tensor model is computed as $dW_h =  \bm b \circ \bm c[h] \circ \bm a$. Both models have the same tensor rank, but the latter has less parameters.}
    \label{fig:diag_lora}
\end{figure*}
\section{Low Rank Tensor adaptation}
\subsection{Parameter sharing across layers}~\label{sec:implicit-tensors}
To further increase the compression ratio in PEFT models, recent works~\citep{2023vera, koohpayegani-nola-2024} suggest sharing parameters across layers that operate as predefined projection matrices. As we see next, this leads to tensor factorization models with fixed parameters.

\textbf{Vector-based Random Matrix Adaptation (VeRA)}
~\cite{2023vera} have proposed to parameterize updates using two learnable vectors at each layer and fixed random matrices shared across all layers. The update at each layer can be expressed as
\begin{align}\label{eq:vera}
 d\tilde{\bm W} = \bm A \Diag\left(\bm C_D[l,:]\right) \bm B^T\Diag\left(\bm C_B[l,:]\right),
\end{align}
where $\bm A, \bm B \in \mathbb{R}^{d\times r}$ are the random projections, and $\bm C_D \in \mathbb{R}^{L\times r}$, $\bm C_B \in \mathbb{R}^{L\times d}$ are matrices that collect trainable vectors across layers. The model in \ref{eq:vera} is a coupled matrix factorization model and is similar to a tensor model. In particular, if we remove the $\bm C_B$ term VeRA can be interpreted as a low-rank tensor CPD parameterization with fixed random factors. That is, the weight update $\tilde{\bm W}$ is a rank-$r$ third order tensor
$\mathcal{T}\in\mathbb{R}^{d\times d \times L}$. Note that, omitting the $C_B$ term has been shown to lead to a small performance degradation unlike omitting $C_D$~\citep{2023vera}.

\textbf{Random Matrix basis Adaptation (NOLA)} In a similar manner,~\cite{koohpayegani-nola-2024} have proposed to parameterize the weight update  by expressing the matrices \(\bm A\) and \(\bm B\) as linear combinations of fixed random basis matrices, that are shared across all layers. The weight update \(d \bm W\) for layer $l$ is then given by:
\begin{equation}\label{eq:nola}
    d\tilde{\bm W}_l = \sum_{i=1}^k \sum_{j=1}^k \alpha_{(i,l)} \beta_{(j,l)} \bm A_i \bm B_j^T, 
\end{equation}
where \(\bm A_i, \bm B_j \in \mathbb{R}^{d \times r}\) are fixed random matrices, shared across all layers, and \(\bm \alpha_l=\left\{\alpha_{(i,l)}\right\}_{i=1}^K\) and  \(\bm \beta_l=\left\{\beta_{(i,l)}\right\}_{i=1}^K\) are the learned coefficients for each layer. If we stack the random matrices \(\bm A_i, \bm B_j \in \mathbb{R}^{d \times r}\) into tensors $\mathcal{A},~\mathcal{B}$ such that: $\mathcal{A}[:,:,i]=\bm A_i$ and $\mathcal{B}[:,:,j]=\bm B_j$, then \ref{eq:nola} can be cast as:
\begin{align*}d\tilde{\bm W}_l & = \sum_{i=1}^k \sum_{j=1}^k \alpha_{(i,l)} \beta_{(j,l)} \sum_{m=1}^r\mathcal{A}[:,m,i] \mathcal{B}[:,m,j]^T \\
    &=\sum_{m=1}^r\mathcal{A}[:,m,:] \left(\bm \alpha_l\bm \beta_l^T\right)\mathcal{B}[:,m,:]^T 
\end{align*}
and $d\tilde{\bm W}_l$ admits the following factorization.
$d\tilde{\bm W}_l = \sum_{m=1}^r{\bm P}_A^{(m)}\left(\bm \alpha_l\bm \beta_l^T\right){\bm P}_B^{(m)T}$, where ${\bm P}_A^{(m)}=\mathcal{A}[:,m,:]$, and ${\bm P}_B^{(m)}=\mathcal{B}[:,m,:]$ are also random projection matrices with different dimensions compared to $\bm A_i,~\bm B_j$. As a result, NOLA can be viewed as the following tensor factorization model:
\begin{align}\label{eq:tensnola}
    d\tilde{\mathcal{W}} = \sum_{m=1}^r \left\llbracket{\bm P}_A^{(m)}{\tilde{\bm A}},{\bm P}_B^{(m)}{\tilde{\bm B}}, {\bm I}\right\rrbracket, \\
    \tilde{\bm A}[:,l]=\bm \alpha_l, \tilde{\bm B}[:,l]=\bm \beta_l.\nonumber
\end{align}
The expression in \ref{eq:tensnola} is a a summation of CPD models, also known as Block Term Decomposition, which is an expressive tensor model, but can lack parsimony~\citep{kolda2009tensor}.
\begin{table*}[hb!]
\centering
\begin{tabular}{llllrr}
\hline
Method & Update Tensor shape & Tensor Model & Parameters & r=4 & r=64 \\
\hline
LoRA & $ML\times d \times d$ & Matrix-Batch & $2MLdr$ & 2.1M & 33M \\
LoReTTA & $ML\times d \times d$ & Custom & $2MLr^2\sum_i k_i$ & 92k & 50M \\
LoTR & $ML\times d \times d$ & Tucker2 & $M(Lr^2+2dr)$ & 33k & 786k \\
FacT-TT & $ML \times d \times d$ & Tensor-Train & $MLr^2+2dr$ & 33k & 786k \\
FacT-TK & $ML \times d \times d$ & Tucker3 & $(2d+ML)r+r^3$ & 33k & 790k \\
Ours & $M\times L\times d\times d/h\times h$ & CP & $(d+d/h+h+L+M)r$ & 17k & 274k \\
\hline
\end{tabular}
\caption{Number of parameters of different low rank PEFT methods as a function of the number of finetuned attention/projection matrices $M$, the number of layers, $L$, the embedding dimension $d$, the number of heads $h$ and the tensor rank of the update, $r$. For LoreTTA,  $k_i$ are hyperparameters that must satisfy $\prod_i k_i = dr$ and $k_i\geq r \;\forall\; i$. We also include the number of parameters for the Llama2-7b architecture when finetuning only M=2 attention matrices (e.g. Q and V) for different ranks. For LoreTTa we use $k_1 = \ldots = k_6=5$ for $r=4$ and $k_1=k_2=k_3=64$ for $r=64$.}\label{tab:peft-params}
\end{table*}

\subsection{LoRTA: A more efficient tensor model}

In the previous section, we explored PEFT models that share parameters across layers, highlighting their correspondence to tensor factorization models. Namely, VeRA and NOLA  utilize fixed projection matrices shared across layers. However, this strategy can result in models that are larger than necessary relative to their degrees of freedom due to the inclusion of these random matrices. Although these matrices can be generated on the fly by solely storing the pseudo-random number generator seed, this still incurs additional resource demands during training, and increases loading time for inference.

To address this issue, we propose modeling the trainable adapters using a low-rank CPD structure. This choice is motivated by the favorable properties of CPD: it is universal, capable of exactly factorizing any tensor, yet remains concise and parsimonious, typically requiring only a small number of parameters to achieve low approximation error~\citep{sidiropoulos2017tensor}. This contrasts with tensor adapters used in vision \citep{2023-tensor-adapters-cnn-aaai} and recently in LLM finetuning~\citep{bershatsky2024lotr}, which rely on Tucker and Tensor-Train models. In fact, for small ranks, CPD is equivalent to Tucker when the core tensor in Tucker is the identity tensor. However Tucker is always parametrized with a dense tensor and therefore requires a larger number of parameters for the same rank.

LoRTA represents all weight updates as a 5th-order tensor $d\tilde{\mathcal{W}}\in\mathbb{R}^{d\times \frac{d}{H} \times H \times L \times M}$. By integrating updates of layers, heads and the \(\bm Q\), \(\bm K\), \(\bm V\), \(\bm P\) matrices into a unified low-rank CPD tensor model, LoRTA exploits redundancy across different modes of the tensor. This approach can thus not only improve parameter efficiency but also facilitate learning by exploiting the shared information among various components of the model. This contrasts with existing PEFT approaches, which tensorize each weight update independently~\citep{yang2024loretta} or only share parameters across layers~\citep{2023-tensor-adapters-cnn-aaai, bershatsky2024lotr}. In order to illustrate how additional tensor modes can result in parameter efficiency gains,   Figure~\ref{fig:diag_lora} compares -- for a single weight update -- LoRA with a rank one tensor model that adds attention heads as a mode.

By utilizing a low-rank CPD model, we can express this tensor as:
\begin{align*}
 d\tilde{\mathcal{W}} = \llbracket \bm{A}, \bm{B}, \bm{C}_H, \bm{C}_L, \bm{C}_M \rrbracket,   
\end{align*}
where \(\bm{A} \in \mathbb{R}^{d \times r}\) and \(\bm{B} \in \mathbb{R}^{\frac{d}{H} \times r}\) are factor matrices for the input and output dimensions, respectively, and \(\bm{C}_H \in \mathbb{R}^{H \times r}\), \(\bm{C}_L \in \mathbb{R}^{L \times r}\), \(\bm{C}_M \in \mathbb{R}^{4 \times r}\) are factor matrices for the attention heads, layers, and the four matrices \(Q\), \(K\), \(V\), \(P\). Each weight matrix update can then be retrieved as:
\begin{align}
& d\tilde{\mathcal{W}}[:, :, k, l, i] = \nonumber \\
& \;\; \bm{A} \left( \text{Diag}\left( \bm{C}_H[k, :] \right) \text{Diag}\left( \bm{C}_L[l, :] \right) \text{Diag}\left( \bm{C}_M[i, :] \right) \right) \bm{B}^\top,\nonumber
\end{align}
where \(k\) indexes the attention heads, \(l\) indexes the layers, and \(i\) indexes the matrices \(\bm Q\), \(\bm K\), \(\bm V\), \(\bm P\). Note that, unlike previous implicit tensor models such as NOLA and VeRA, which rely on fixed random projections or parameters and learn only scaling coefficients, our proposed model is \textit{fully trainable}. All factor matrices (\(\bm{A}\), \(\bm{B}\), \(\bm{C}_H\), \(\bm{C}_L\), \(\bm{C}_M\)) are learned during training, providing greater expressiveness and forgoing the dependency on pre-defined random bases or projections.

Table~\ref{tab:peft-params} shows how the CP low rank tensor parameterization leads to better scaling in the number of parameters with respect to the tensor rank $r$. Moreover, our higher-order weight update tensorization improves scaling in terms of transformer architecture hyperparameters, namely the embedding dimension $d$, number of attention heads $H$, and number of fine-tuned attention matrices $M$.

\subsection{Other Low Rank Tensor models in PEFT}

As mentioned in the previous section, existing PEFT tensor-based models differ from our method both in their parameter-sharing schemes, which result from different weight update tensorization approaches, as well as in the low-rank tensor models they employ. Below, we provide a concise overview of these approaches which intends to highlight the differences with LoRTA; further details are available in Appendix~\ref{a:sec:other-models} and the provided references. 

\textbf{FaCT \& LoTR} In the context of vision transformers,~\cite{2023-tensor-adapters-cnn-aaai} have proposed to represent updates across all layers as a \emph{third} order tensor $d\tilde{\mathcal{W}}\in\mathbb{R}^{L \times d\times d }$. They propose two parameterizations of $d\tilde{\mathcal{W}}$, namely, a Tensor Train and Tucker3 low rank tensor models. 
Recently,~\cite{bershatsky2024lotr} have proposed to apply the same tensorization across layers to fine-tune LLMs, but using a low rank Tucker2 tensor model to parameterize updates:
\begin{align}
d\tilde{\mathcal{W}} = ;; \bm{G} \times_1 \bm{A} \times_2 \bm{B}
\end{align}
where $\bm{A}$, $\bm{B}$ $ \in \mathbb{R}^{d\times r}$ and $\bm{G} \in \mathbb{R}^{L\times r\times r}$.

\textbf{LoreTTA}~\cite{yang2024loretta} propose two methods that employ low rank tensor models. However, these models do not share parameters across layers, they reparameterize low rank matrix adapters using low rank tensor models. In \emph{LoreTTA-rep} a low rank matrix model is first applied to each weight update in the same manner as described for LoRA in Equation~\eqref{eq:lora}. Then each of the $ML$ resulting LoRA factors $\bm A, \bm B \in \mathbb{R}^{d\times r}$ are expressed as a n-th order tensor with arbitrary dimensions, i.e. $\mathcal{A}, \mathcal{B} \in \mathbb{R}^{k_1 \times \ldots\times k_N}$.    Finally, each of these tensors is parametrized  Tensor Train model, explicitly, 
\begin{align}
\mathcal{A} = \prod_{i=1} \bm G_i \quad \text{ where } \;\bm G_i \in \mathbb{R}^{r\times k_i\times r}.
\end{align}
We highlight that the added dimensions $k_i$ are hyperparameters that must satisfy $\prod_i k_i = dr$ and $k_i\geq r \;\forall\; i$; otherwise, it would induce a new tensor rank deficiency. 
Moreover, choosing appropriate values of $k_i$ might be challenging and necesitate further hyperparameter tuning. \cite{yang2024loretta} also proposed \emph{LoReTTA-adp}, applying a tucker parameterization to an adapter method, which unlike our method and the rest of the aforementioned methods adds new parameters to the model and thus can not be merged into the original weights, thereby incurring additional inference costs. 

\section{Experiments}\label{sec:experiments}
\begin{table*}[ht!]
\setlength{\tabcolsep}{5pt}
\centering
\begin{tabular}{l | r | ccccccc | c}
\toprule
 & Method & \parbox{1.5cm}{\footnotesize\# Trainable \\ Parameters} & SST-2 & MRPC & CoLA & QNLI & RTE & STS-B & Avg. \\
\midrule
\multirow{4}{*}{\rotatebox{90}{LoReTTA}} 
& LoRA (r=8) & 630k & 94.01 & 91.48 & 62.08 & 92.39 & 74.51 & 84.69 & 83.19 \\
& LoReTTA rep & 70k & 94.28 & 90.63 & 61.72 & 92.40 & 74.42 & 89.24 & 83.78 \\
& LoRTA (r=20) & 48k & 94.27 & 92.04 & 63.35 & 91.48 & 75.09 & 89.82 & \textbf{84.34} \\
& LoRTA (r=12) & 29k & 93.81 & 91.13 & 61.40 & 92.04 & 74.73 & 89.64 & 83.79 \\
\midrule
\multirow{4}{*}{\rotatebox{90}{LoTR}} 
& LoRA (r=8) & 300k & 94.2 & 88.0 & 61.1 & 91.3 & 73.0 & 90.7 & 83.05 \\
& LoTR & 74k & 93.0 & 85.9 & 60.5 & 90.0 & 66.0 & 91.9 & 81.22 \\
& LoRTA (r=16) & 15k & 94.73 & 90.44 & 64.32 & 92.37 & 76.9 & 90.25 & \textbf{84.84} \\
& LoRTA (r=4) & \textbf{3.4k} & 94.61 & 89.21 & 60.55 & 90.61 & 76.9 & 89.97 & 83.6 \\
\midrule
\multirow{4}{*}{\rotatebox{90}{VeRA}} 
& LoRA & 800k & 96.2 & 90.2 & 68.2 & \textbf{94.8} & 85.2 & \textbf{92.3} & \textbf{87.8} \\ 
& LoRA-FA & 3.7M & 96.0 & 90.0 & 68.0 & 94.4 & 86.1 & 92.0 & 87.7 \\
& VeRA & 61k & 96.1 & \textbf{90.9} & 68.0 & 94.4 & \textbf{85.9} & 91.7 & \textbf{87.8} \\ 
& LoRTA (r=8) & \textbf{9k} & 96.3 & 89.5 & 65.1 & 94.3 & 85.6 & 91.1 & 85.7 \\
\bottomrule
\end{tabular}
\caption{Performance of RoBERTa Base and Large models on GLUE tasks under three different experimental settings reported by LoReTTA~\citep{yang2024loretta}, LoTR~\citep{bershatsky2024lotr}, and VeRA~\citep{2023vera}. In LoReTTA, LoRTA is applied to the encoder and LoRA to the classifier with the same rank, while for LoTR and VeRA, LoRTA is applied only to the encoder. Trainable parameters include the classifier for LoReTTA but exclude it for LoTR and VeRA, where it is fully trained. VeRA results use RoBERTa Large, whereas LoTR and LoReTTA use RoBERTa Base.}
\label{tab:glue-consolidated}
\end{table*}

\subsection{Natural Language Understanding}

We evaluate our approach by fine-tuning RoBERTa models on the General Language Understanding Evaluation (GLUE)~\cite{wang2018glue} benchmark. We conduct experiments across three distinct settings previously reported in the literature by~\cite{bershatsky2024lotr}, ~\cite{yang2024loretta} and \cite{2023vera}. These settings differ in hyperparameters, including the number of training epochs,  different learning rates for the classification head and encoder, the learning rate decay strategy (linear vs fixed), the use of different scaling parameters $\alpha$, and the grid search ranges. Because the best results on the validation set across a grid of hyperparameter values are reported, performance for the same baseline method can vary considerably across settings (see, for example, LoRA performance reported by \cite{hu2021lora}, \cite{yang2024loretta} and \cite{bershatsky2024lotr}). Therefore, we provide an evaluation of our method in a variety of experimental conditions, while also maintaining the original configurations in which state-of-the-art methods were originally reported. Detailed settings can be found in Appendix~\ref{a:sec:hparams:nlu}.

We also finetuned Llama2 models~\citep{touvron2023llama} on question-answering (QA) tasks SQuAD~\citep{rajpurkar2016squad}, DROP~\citep{dua2019drop}, COPA~\citep{copa}, and ReCoRD~\citep{zhang2018record}, following the experimental setting outlined by~\cite{yang2024loretta}. For these tasks, we used a randomly selected subset of 1,000 samples to simulate a low-data regime and increase the task difficulty. All classification tasks are tackled as language modeling tasks following the prompt-based fine-tuning approach described by~\cite{Malladi2023FinetuningLM}.

\textbf{Baselines}
We benchmark our method against the following methods:
\begin{itemize}
    \item \textbf{Full finetuning}:  all parameters are trained.
    \item \textbf{IA3}~\citep{Liu2022FewShotPFIA3}: rescales activations with learned
vectors
    \item \textbf{Prefix}~\citep{Li2021PrefixTuning}: prepends learnable continuous vectors (prefixes) to the input embeddings.
    \item \textbf{LoRA}~\citep{hu2021lora}, \textbf{LoRA-FA}~\citep{zhang2023lorafa} and \textbf{VeRA}~\citep{2023vera}, \textbf{LoTR}~\citep{bershatsky2024lotr}, \textbf{LoReTTA-rep}~\citep{yang2024loretta}: As previously described.
    \item We omit \textbf{Adapter$^{\mathbf{H}}$}~\citep{houlsby2019parameter}, \textbf{Adapter$^{\mathbf{P}}$}~\citep{pfeiffer2020adapterfusion}, \textbf{Bitfit}~\citep{zaken2021bitfit}, \textbf{AdapterDrop}~\citep{ruckle2020adapterdrop}, and other methods that are customarily reported in PEFT literature but have been outperformed by more recent methods in these settings.
\end{itemize}

The results in Table~\ref{tab:glue-consolidated} show that LoRTA can achieve comparable or slightly superior performance with less trainable parameters compared to state of the art tensor based PEFT methods LoreTTA~\citep{yang2024loretta} and LoTR~\citep{bershatsky2024lotr} when finetuning RoBERTA base on GLUE tasks. Similarly, for RoBERTa large LoRTA can also achieve a 6x reduction in the number of trainable parameters with only small drop in average performance (2\%) when compared to ~\cite{2023vera}. In this settings we did not tune the hyperparameters for our method as extensively as baselines, and thus this gap could be further reduced. 

In Llama QA experiments, shown in Table~\ref{tab:llama-qa}, full fine-tuning (FT) achieves the highest average score (77.3) with 7 billion trainable parameters, but  among the PEFT methods LoRTA (r=8) achieves the highest average score (76.7) with just 0.03 million parameters, representing a 17x reduction in parameter count with respect to the most efficient method. 

\begin{table*}[ht!]
\centering
\begin{tabular}{l | r | cccc | c}
\toprule
Method & \parbox{2cm}{\footnotesize\# Trainable \\ Parameters} & COPA & ReCoRD & SQuAD & DROP & Avg. \\
\midrule
Full & 7B & 86 & 81.1 & 90.71 & 51.38 & 77.3 \\
LoRA (r=8) & 4.19M & 81 & 79.4 & 90.56 & 45.96 & 74.2 \\
Prefix & 1.31M & 83 & 81.0 & 90.56 & 45.95 & 75.1 \\
IA3 & 0.60M & 80 & 81.5 & 89.41 & 39.37 & 72.6 \\
LoRETTA rep & 0.51M & 86 & 80.3 & 88.47 & 42.71 & 74.4 \\
LoRTA (r=4) & \textbf{0.02M} & 87 & 81.1 & 87.4 & 44.04 & 74.9 \\
LoRTA (r=8) & \textbf{0.03M} & 87 & 81.6 & 88.5 & 49.7 & \textbf{76.7} \\
\bottomrule
\end{tabular}
\caption{Llama2-7B performance on SuperGLUE and question-answering tasks (SQuAD, DROP). We follow the experimental setup used by~\cite{yang2024loretta}.}
\label{tab:llama-qa}
\end{table*}

\subsection{Instruction Tuning}

We fine-tune the 7 billion parameter Llama2~\citep{touvron2023llama} models on the cleaned Alpaca instruction tuning dataset~\citep{alpaca}. While more recent models and tasks exist, we select this well-studied setting because it enables direct comparison with an extensive body of prior work, and maintains methodological consistency with our earlier experiments on NLU tasks.  We train for one epoch, preceded by a warm-up learning rate sweep as in the standard setting. Other hyperparameters are detailed in Appendix~\ref{a:sec:hparams:inst}.

As shown in Figure~\ref{fig:loss-llama-7b}, LoRTA effectively reduces the number of parameters to a fraction of those required by the lowest rank in LoRA, with only a small performance cost. In this setting the validation cross entropy decreases monotonically with the number of parameters used, both in training and testing, and LoRTA even demonstrates superior performance with fewer parameters for ranks 96 and 192.
\begin{figure*}[hb!]
\centering
\includegraphics[width=0.8\textwidth]{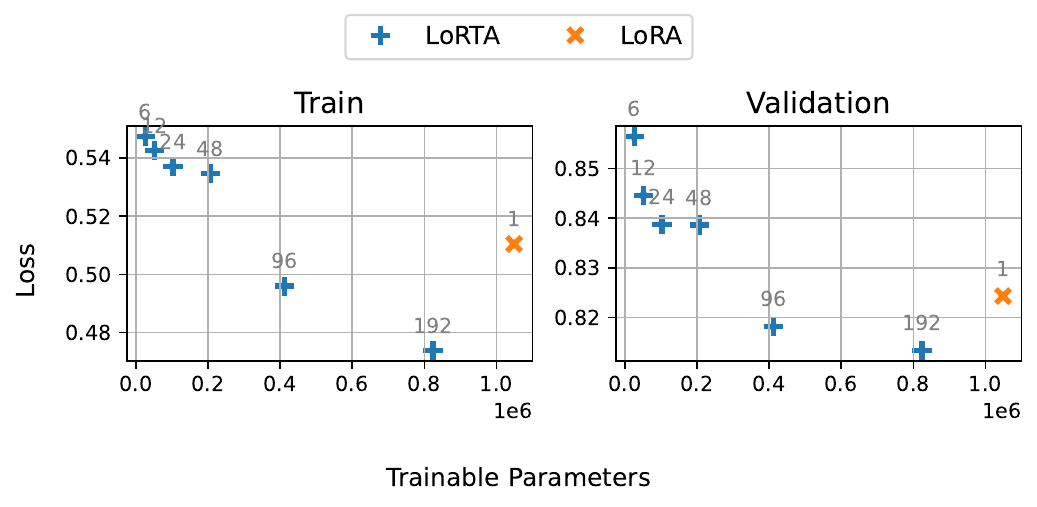}
    \caption{Mean cross-entropy loss on training and testing data for Llama2-7b on the Alpaca dataset vs number of trainable parameters for different adapter ranks. Lower is better. Numbers on top of markers denote the adapter rank.}
    \label{fig:loss-llama-7b}
\end{figure*}

Although cross entropy (and perplexity) has been shown to be correlated with diverse downstream performance metrics~\cite{dubois2024lengthcontrolledalpacaevalsimpleway}, we provide additional evaluations using other standard LLM-as-a-judge benchmarks in Appendix~\ref{a:sec:results}, which also show LoRTA attains comparable performance to LoRA at a fraction of parameters.

\subsection{Preference Optimization}

Among the various existing techniques to align LLMs with human preferences on specific tasks--see, for example,~\cite{kaufmann2023survey-rlhf} and references therein-- we utilize Direct Preference Optimization (DPO)~\citep{rafailov2024dpo} due to its widespread use. We set the regularization coefficient that penalizes deviations from the pre-trained model's outputs ($\beta$) to $0.1$. We use the cleaned version of the Intel Orca dpo pairs dataset\footnote{\texttt{https://huggingface.co/datasets/\\argilla/distilabel-intel-orca-dpo-pairs}}. 
 We use Huggingface Transformer Reinforcement Learning
(trl) library\footnote{\texttt{https://github.com/huggingface/trl}}. Consistent with our previous experiments, we use Llama2-7b as our base model. For a complete description of hyperparameters see Appendix~\ref{a:sec:hparams:dpo}.
\begin{table}[h]
    \centering
    \begin{tabular}{c| c|c|c}
        \toprule
          &\# Parameters & Val. Loss & \parbox{2.0cm}{MT-bench \\ Score} \\
        \hline
         LoRA   & 524 k & 0.44 & 4.08 \\
         LoRTA  & 4 k   & 0.43 & 4.14  \\
        \bottomrule
    \end{tabular}
    \caption{Llama2-7b fineuned using DPO on the orca dataset. Both methods use rank 1. We report the DPO loss for held-out data (lower is better) and the average score across MT-bench tasks.}
    \label{tab:DPO}
\end{table}

The results in Table~\ref{tab:DPO} compare both methods using rank 1 updates. LoRTA achieves comprable performance to LoRA in terms of validation loss, while using only 4k parameters—a 99\% reduction from LoRA's 524k parameters. In addition, LoRTA attained a slight improvement in MT-Bench~\cite{mtbench} performance, showing differences in generalization across tasks. Additional results for different ranks can be found on Appendix~\ref{a:sec:results}. Unlike instruction tuning, preference across ranks exhibits non-monotonic behaviour, and larger ranks do not lead to performance improvements.

\subsection{Protein Folding}
Protein folding, the process by which a protein's amino acid sequence determines its three-dimensional structure, is a fundamental problem in molecular biology. Accurate prediction of protein structures from their sequences has significant implications for understanding protein function and designing new proteins for therapeutic purposes.  ESMFold \citep{Lin2021ESMFold} is a frontier model for this task trained in two stages. First, ESM-2, a BERT-based \citep{devlin2019bertpretrainingdeepbidirectional} protein language model, is trained with the masked-language-modeling objective on amino acid sequences. This unsupervised pretraining allows the model to capture complex patterns and relationships within protein sequences. Remarkably, valuable structural information emerges in the model's features during this process \citep{Rao2020.12.15.422761}. In the second stage, ESM-2 is frozen, and a model head predicting three-dimensional protein structures is trained on top of language model features. 

We re-train ESMFold in the second stage – fine-tuning ESM-2 parameters (we use ESM-2 35M instead of the ESM-2 3B model used in \cite{Lin2021ESMFold} due to compute constraints) with LoRA and LoRTA instead of freezing them. We evaluate performance with the Local Distance Difference Test for C$\alpha$ atoms (LDDT-C$\alpha$) \citep{Mariani2013lDDT} – that measures accuracy of predicted protein structures by comparing the distance between alpha carbons in predicted and true structures. LDDT-C$\alpha$ ranges from 0 (poor accuracy) to 1 (perfect match). See Appendix~\ref{a:sec:hparams:folding} for experiment details. 

\begin{table}[h]
    \centering
    \begin{tabular}{c|c|c|c}
        \toprule
         & Rank &\parbox{1.5cm}{\footnotesize\# Trainable \\ Parameters \\ ($\times 10^4$)} & LDDT-Ca \\
        \hline
         LoRA   & 1 & 3.5 & 0.67 \\
         \hline
        \multirow{2}{*}{LoRTA} & 1  & 0.1 & 0.66 \\
        & 8 & 0.5 & 0.67 \\
        \bottomrule
    \end{tabular}
    \caption{Mean LDDT-C$\alpha$ on held-out data. Higher is better. LoRTA rank 1 is competitive with LoRA rank 1 on the test set despite having 64x fewer parameters.}
    \label{tab:folding}
\end{table}

As shown in Table~\ref{tab:folding}, LoRTA with ranks 1 and 8 achieves performance comparable to LoRA rank 1, despite using an order of magnitude fewer parameters. In Appendix~\ref{a:sec:results}, we report training losses and results for larger ranks. Notably, larger ranks for LoRTA do not improve performance.

\subsection{Computational Advantages}

\textbf{Reduced I/O times for concurrent adapters.}
During training, the GPU memory usage and training times  of most low rank methods are comparable.
Nonetheless, further reductions in adapter size are motivated by (i) lower storage requirements (ii) the need to improve task-switching efficiency and minimize storage requirements in scenarios involving a large number—potentially thousands—of adapters. Frequent CPU-GPU transfers for loading adapters in such settings can introduce significant overhead. By further compressing parameters, it becomes feasible to load several customized models with a shared base LLM in GPU memory, substantially enhancing scalability and performance in multi-task environments.

\begin{table}[ht!]
\setlength{\tabcolsep}{3pt}
\centering
\begin{tabular}{l|r|rrr}
\toprule
r & n & \multicolumn{3}{c}{Transfer time (ms)} \\
& & \multicolumn{1}{c}{Lora} & \multicolumn{1}{c}{Lotr} & \multicolumn{1}{c}{Lorta} \\
\midrule
\multirow{3}{*}{4} 
& 1 & 10.0\,(9.0) & 0.4\,(0.2) & \textbf{0.12}\,(0.03) \\
& 10 & 12.64\,(0.08) & 3.25\,(0.05) & \textbf{1.06}\,(0.01) \\
& 100 & 144.0\,(2.0) & 17.1\,(0.2) & \textbf{5.5}\,(0.2) \\
\midrule
\multirow{3}{*}{64} 
& 1 & 23.0\,(5.0) & 3.4\,(0.2) & \textbf{1.08}\,(0.01) \\
& 10 & 216.0\,(2.0) & 41.02\,(0.04) & \textbf{8.811}\,(0.005) \\
& 100 & 2272\,(31) & 375.1\,(0.2) & \textbf{61.5}\,(0.1) \\
\bottomrule
\end{tabular}
\caption{CPU to GPU transfer time in milliseconds (mean (std) across 20 repetitions) of $N$ concurrent adapters with rank r for Llama2-7b using an NVIDIA A6000. Boldface indicates smallest value within one standard deviation.}
\label{tab:gpu-transfer-times}
\end{table}
To illustrate the potential latency reduction obtained from more compact adapters, in Table~\ref{tab:gpu-transfer-times} we present CPU to GPU transfer times for different adapter methods with the same rank. Additional hardware profiling results for memory usage and training time can be found on Appendix~\ref{a:sec:profiling}.

\textbf{Tensor Decomposition of trained adapters is challenging.}
We also explore obtaining similarly compact representations by decomposing pre-trained LoRA adapters. To empirically demonstrate the challenges of post-training tensor decomposition, we conducted experiments using trained LoRA adapters from our preference optimization task (using Llama2-7b as a base model, and rank 8 for the adapters). Using TensorLy's\footnote{\texttt{https://tensorly.org}} implementation of CP decomposition via alternating least squares (ALS), we attempted to decompose each weight matrix $W \in \mathbb{R}^{d \times d}$ into a rank-8 CP model by reshaping it into a 3rd-order tensor $\mathcal{T} \in \mathbb{R}^{d \times \frac{d}{h} \times h}$, where $h$ is the number of attention heads. This tensorization scheme limits our proposed LoRTA model's decomposition to attention heads, i.e., without the additional structure across layers and matrices.

\begin{table}[h]
\centering
\begin{tabular}{l|cccc}
\toprule
Metric & Mean & Median & Max & Std \\
\midrule
Relative Error & 0.827 & 0.838 & 0.910 & 0.053 \\
$R^2$          & 0.312 & 0.297 & 0.506 & 0.086 \\
\bottomrule
\end{tabular}
\caption{Approximation quality metrics for CP decomposition of trained LoRA weights across all layers and Q/V matrices. The high relative error and low $R^2$ scores indicate poor reconstruction quality.}
\label{tab:decomp-stats}
\end{table}

The results in Table~\ref{tab:decomp-stats} reveal remarkably high approximation errors that render the compressed adapters ineffective, even when targeting the same parameter count achieved by direct tensor-based training. This suggests incorporating low rank tensor structure during training can guide the optimization toward fundamentally different, more compressible parameter updates. This finding parallels similar observations in neural network compression~\citep{hoefler2021sparsity}, where incorporating structural constraints during training often yields better results than post-hoc pruning.

\section{Conclusion}
We have introduced LoRTA, a novel approach that employs a low-rank tensor model for LLM updates. By extending low-rank adaptation to higher-order tensors, LoRTA overcomes the inherent lower bounds on the number of trainable parameters while offering finer-grained control over adapter size. Our experiments across various benchmarks demonstrate that LoRTA achieves comparable and sometimes superior performance than baselines at a reduced parameter count.
\FloatBarrier
\section{Limitations and Future Work}

While our experiments demonstrate the LoRTA can be used to finetune models effiently across various settings comprising different tasks and model architectures, there are several important limitations and directions for future empirical research on the proposed adaptation method. 

First, we have shown that previous works have implicitly utilized low-rank tensor models with random factors. Nothing precludes our higher-order tensor model from using randomized factors for increased efficiency—a potential direction for future work that could further reduce computational overhead. Lastly, developing more efficient implementations of  tensor operations that result in greater memory efficiency also remains a relevant future work direction which could make LoRTA even more suitable for resource-constrained environments.

Second, our evaluation was constrained to models up to 7B parameters due to computational limitations, with LLaMA 2 being the latest model tested. Further research is needed to assess LoRTA's scalability and effectiveness on larger models (e.g., 70B+ parameters) and more recent architectures. Additionally, understanding how parameter efficiency gains evolve with increasing model size remains an open question. Expanding our evaluation beyond standard benchmarks to multimodal models, text-to-speech systems, and domain-specific adaptation scenarios could provide deeper insights into the method's generalizability and robustness. Moreover, our study did not incorporate human evaluations, which could offer more nuanced assessments of LoRTA’s impact on model quality and usability.

While our method shows promise for concurrent adapter scenarios, further research is needed to evaluate its effectiveness in these settings, including adapter composition, cross-task transfer, and adapter merging. Additionally, exploring LoRTA in the context of Mixture of Experts (MoE) architectures—where experts could be parameterized as tensor factors—represents an interesting direction that could enhance both parameter and computational efficiency. The potential for sharing tensor factors across experts or dynamically adjusting tensor ranks based on task complexity remains unexplored.

The current tensorization scheme, while effective, represents just one possible approach. Alternative schemes might offer different efficiency-performance trade-offs or be better suited for specific architectures or tasks. For instance, incorporating additional modes based on model-specific features (like relative position embeddings or sliding window attention patterns) could potentially yield further improvements. Moreover, our method currently focuses on attention matrix adaptation, and extending it to other components like MLPs or embeddings warrants investigation.

Further empirical investigation could provide valuable insights through ablation studies on the impact of tensor rank across different settings, detailed analysis of the learned tensor factors, and examination of how different tensorization schemes affect various aspects of model behavior.

Our work addresses only the parameter-efficient fine-tuning aspect of model adaptation. Future research could explore combining LoRTA with other efficiency techniques such as quantization, pruning, or activation compression. Additionally, while we demonstrated improved I/O efficiency for concurrent adapters, developing more efficient implementations of tensor operations could further reduce memory usage and training time. This includes leveraging hardware-specific optimizations and exploring methods to compress or efficiently compute intermediate activations.

From a theoretical perspective, several questions remain open. Understanding why tensor-based adapters provide an effective inductive bias for model adaptation, and characterizing what different adapter architectures learn, could provide insights for designing better adaptation methods. Additionally, while we focused on CP decomposition due to its parameter efficiency, comparative studies with other tensor decompositions (e.g., Tucker, Tensor Train) could reveal interesting trade-offs between expressivity and efficiency.  Finally, while our preliminary experiments suggest that incorporating low-rank structure during training leads to more compressible updates than post-hoc decomposition, a deeper understanding of this phenomenon could inform the development of improved adaptation methods.

\bibliography{references}

\onecolumn
\clearpage  
\appendix
\twocolumn
%
\section{Tensor Algebra}\label{App:tensors}

To facilitate our analysis, we briefly present some tensor algebra preliminaries and refer the reader to \cite{sidiropoulos2017tensor,kolda2009tensor} for further details.
	
A $N$-order tensor $\mathcal{\bm X}\in\mathbb{R}^{I_1\times I_2\times\dots\times I_N}$ is an $N$-way array indexed by $i_1,i_2,\dots,i_N$ with elements $\mathcal{\bm X}(i_1,i_2,\dots,i_N)$. 
It consists of $N$ types of modes: $\mathcal{\bm X}(:,i_2,\dots,i_N)$, $\mathcal{\bm X}(i_1,:,\dots,i_N),\dots,\mathcal{\bm X}(i_1,i_2,\dots,:)$.
	
	A rank-one tensor $\mathcal{\bm Z}\in\mathbb{R}^{I_1\times I_2\times\dots\times I_N}$ is the outer product of $N$ vectors defined as:
 \begin{equation}
     \mathcal{\bm Z}=\bm a_1\circ \bm a_2\circ\dots\circ \bm a_N,
 \end{equation}
 where $\bm a_1\in\mathbb{R}^{I_1},~\bm a_2 \in\mathbb{R}^{I_2},\dots,~\bm a_N \in\mathbb{R}^{I_N}$ and $\circ$ denotes the outer product. The elementwise formula of the above expression is:
	\begin{equation}
	    \mathcal{\bm Z}(i_1,i_2,\dots,i_N)=\bm a_1(i_1)\bm a_2(i_2)\cdots\bm a_N(i_N),~~ \forall i_1,i_2,\dots,i_N,
	\end{equation}
 Any tensor can be realized as a sum of $N$-way outer products (rank one tensors), i.e. 
	\begin{equation}\label{PD-app}
	\mathcal{\bm X}=\sum_{r=1}^R\bm a_1^{f}\circ \bm a_2^{f}\circ\dots\circ \bm a_N^{f}.
	\end{equation}
	The above expression represents the \textit{canonical polyadic decomposition} (CPD) or \textit{parallel factor analysis} (PARAFAC)~\citep{harshman1994parafac} of a tensor. The CPD elementwise representation is:
	\begin{equation}
	    \mathcal{\bm X}(i,j,k)=\sum_{r=1}^R{\bm A_1}(i_1,f){\bm A_2}(i_2,f)\cdots{\bm A_N}(i_N,f),
	\end{equation}
	 where $\bm{A_n}=[\bm a_n^1,\bm a_n^2,\dots,\bm a_n^F]\in\mathbb{R}^{I_n\times F},~n=1,\dots,N$ are called the low rank factors of the tensor. A tensor can be fully characterized by its latent factors, so we can represent a tensor by its CPD model as:
  \begin{equation}
      \mathcal{\bm X} =\left\llbracket{\bm A_1},{\bm A_2},\dots,{\bm A_N}\right\rrbracket.
  \end{equation}

A tensor can be also represented as a set of matrices, by fixing all the modes but two as:
  \begin{align}
      &\mathcal{\bm X}\left[:,:,i_3,\dots,i_N\right]=\nonumber\\&{\bm A_1}\left(\Diag\left(\bm A_3\left(i_3,:\right)\right)\odot\cdots\odot\Diag\left(\bm A_N\left(i_N,:\right)\right)\right){\bm A_2^T},
  \end{align}  
where $\Diag\left(\bm A_n\left(i_n,:\right)\right)$ is the diagonal matrix with diagonal equal to $\bm A_N\left(i_n,:\right)$.

\section{Additional related work}\label{a:sec:add-related}

\textbf{Efficient Architectures}  Another relevant direction in reducing resource usage is using more efficient model architectures. Mixture of Experts (MoE) technique, implemented in models like Switch Transformers~\citep{fedus2022switch} and GLaM~\citep{pmlr-v162-du22c-glam}, has shown promise in scaling model capacity while maintaining computational efficiency by activating only relevant sub-models for given inputs. Recent works~\cite{buehler2024xlora, zhang2024milora} have explored parameterizing experts, which often amount to different feed forward module parameters within the transformer block, using low rank adapters. modules\cite{bershatsky2024lotr} have proposed that experts in Mixture of Experts (MoE) models could be also modeled jointly as a fourth order tensor $d\tilde{\mathcal{W}}_m\in\mathbb{R}^{d\times d\times L\times E}$, where $E$ is the number of experts, but no tensor based models were explored in practice. There is also relevant work on non-transformer architectures, such as RWKV~\citep{peng2023rwkv} and Mamba~\citep{gu2023mamba}. PEFT methods for these architectures have also been explored~\cite{mamba2025peft, ham2024prodial}.

\textbf{Model Compression} While these techniques differ from PEFT in that they focus on reducing the requirements of a trained model rather than efficient adaptation, they offer valuable insights for developing more efficient PEFT approaches. Pruning and quantization are key techniques for compressing neural networks, that have also been extensively applied to LLMs. Pruning removes less important weights, with some methods achieving high compression rates, e.g.~\citep{ma2023llmprunning}. Quantization reduces weight precision, decreasing model size and also allowing more efficient operations~\citep{lin2024awq}. The combination application of PEFT methods with quantization or pruning techniques to further improve efficiency has been explored, for example in~\cite{dettmers2024qlora, benedek2024prilora}.

\textbf{Data efficient fine tuning.} An alternative approach to reducing fine-tuning costs is to reduce the amount of data. In this direction, Few-shot and continual learning approaches have been shown to be effective in LLM fine-tuning tasks~\citep{lin2024dataeff, wang2024inscldataeff}.
%
%

\textbf{Low Rank Training.} Exploiting low rank structure to improve efficiency during both training and inference in deep models has long been studied~\citep{sainath2013low}, and also combined with sparsity~\citep{sprechmann2015learning}. Recent advancements include Cuttlefish~\citep{wang2023cuttlefish} and ELRT~\citep{sui2024elrt}. 

\section{Other Tensor Low Rank Models in PEFT}\label{a:sec:other-models}

\subsection{Tensorizing individual weight updates.}
~\cite{yang2024loretta} propose to each low rank matrix in a LoRA adapter. Explicitly, if for a single weight update  $d\tilde{\bm W}\in\mathbb{R}^{d\times d}$, is first expressed using the low rank matrix model
\begin{align}\label{eq:lora}
 d\tilde{\bm W} = \frac{\alpha}{r}\bm A \bm B^T, \quad \bm A, \bm B \in \mathbb{R}^{d\times r},
\end{align}
Then, both low rank matrix factors $\bm A$ and $\bm B$ are expressed as order-$D$ tensors $\mathcal{A}, \mathcal{B} \;\in \mathbb{R}^{k_1 \times \ldots \times k_D}$, where the chosen dimensions must satisfy $\prod_{i=1}^D k_i = d^2$.
The condition $k_i\geq r \;\forall\; i$ also imposes a limit on the order of this tensor: $n\leq \log_r(d)$. Since $n>2$ is required for LoreTTA to be potentially more efficient than LoRA, $r>\sqrt{d}$, which This can be limiting in practice since $n > 2$ is required for LoReTTA to be potentially more efficient than LoRA. This implies that $r > \sqrt{d}$, which can be limiting in practice. For example, $r \leq 64$ for a Llama-7B model.

These tensorized updates are then parameterized using a low rank Tensor-Train model with equal ranks across all factors. Explicitly:
\begin{align*}
   d\tilde{\mathcal{A}} =  \prod_{i=1}^D \mathcal{G}_i, \quad \mathcal{G}_i \in \mathcal{R}^{r\times k_i\times r}.
\end{align*}

\subsection{Parameter sharing across layers}
Both~\cite{2023-tensor-adapters-cnn-aaai} in the context of vision transformers and \cite{bershatsky2024lotr} for LLMs have proposed to represent the updates of each fine-tuned attention matrix ($\bm{Q}, \bm{K}, \bm{V}, \bm{P}$) across all layers as a tensor $d\tilde{\mathcal{W}}_m\in\mathbb{R}^{d\times d\times L}$. \cite{bershatsky2024lotr} parametrize it using a Tucker-2 model
\begin{align}
d\tilde{\mathcal{W}}_m = \;\; \bm{G} \times_1 \bm{A} \times_2 \bm{B} ,
\end{align}
where $\bm{A}$, $\bm{B}$ $ \in \mathbb{R}^{d\times r}$ and $\bm{G} \in \mathbb{R}^{L\times r\times r}$.

\section{Parameter efficiency gains breakdown.}\label{a:sec:breakdown}

We provide a breakdown of the parameter savings achieved by our proposed method, LoRTA, compared to LoRA, by parameterizing the weight updates using low-rank tensor decompositions at different granularities -- i.e., shared modes--. Table~\ref{a:tab:breakdown} summarizes the dimensions of the update tensors, the number of update tensors used, and the corresponding parameter savings when the tensor rank $r$ matches the tensor rank of LoRA rank $r$. The first row corresponds to LoRA.

To fairly compare the parameter efficiency of LoRTA with LoRA, we adjust the tensor rank in LoRTA to match the effective total tensor rank in LoRA, which is $r^{\prime}=r \times 4 L$ due to LoRA applying a rank $r$ update to each of the $4 L$ matrices individually. For a given tensor rank, LoRTA reduces the number of parameters from scaling  $8 d L r$ in LoRA to $4L(d(1+1/h)+h+L+4)r$ in LoRTA (usually $d\gg L$ and $d\gg h$), achieving substantial parameter savings without compromising expressive power. For example, this amounts to a $47.6 \%$ reduction in a LLaMA2 7B model.

 \begin{table*}[h!]
\setlength{\tabcolsep}{5pt}
\centering
\caption{Breakdown of the parameter savings against LoRA by mode, i.e., parameter sharing dimension. This is the number of parameters required to represent an update with thensor rank $r$.}\label{a:tab:breakdown}
\begin{tabular}{l | c | c | c}
\toprule
\textbf{Added Modes} & \textbf{Update Tensor Dimensions} & \textbf{Number of Update Tensors} & \textbf{Parameter Savings} \\ 
\midrule
 & $d \times d$ & $4L$ & $0$ \\ 
Heads & $d \times \frac{d}{H} \times H$ & $4L$ & $1-\frac{d\left(1 + \frac{1}{H}\right) + H}{2dr}$ \\ 
Heads, QKVP & $d \times \frac{d}{H} \times H \times 4$ & $L$ & $1-\frac{d\left(1 + \frac{1}{H}\right) + H + 4}{2dr}$ \\ 
Heads, QKVP, Layers & $d \times \frac{d}{H} \times H \times 4 \times L$ & $1$ & $1-\frac{d\left(1 + \frac{1}{H}\right) + H + 4 + L}{2dr}$ \\ 
\bottomrule
\end{tabular}
\end{table*}

\section{Experimental details}\label{a:sec:hparams}
In this appendix, we provide further details on the experiments presented in the main paper.
\subsection{NLU}\label{a:sec:hparams:nlu}
In our GLUE experiments we implemented our method using Huggingface's PEFT,  VeRA~\cite{2023vera} and LoreTTA~\cite{yang2024loretta} codebases. Hyperparameters for each of the three settings reported are
detailed below.
\begin{table}[h!]
\setlength{\tabcolsep}{5pt}
\centering
\label{tab:model_hyperparameter_config}
\begin{tabular}{l | c}
\toprule
\textbf{Hyperparameter} & \textbf{Value} \\
\midrule
$\alpha$ & 16 \\
Learning Rate & [2E-3, 5E-4] \\
Scheduler & Constant \\
Optimizer & AdamW \\
Number of Epochs & 20 \\
Batch Size & [16, 32] \\
Warmup Steps & 500 \\
\bottomrule
\end{tabular}
\caption{Hyperparameter configurations for RoBERTa Base on the GLUE benchmark following the setup reported by~\cite{yang2024loretta}, where only the batch size and learning rate are tuned for each task, selecting between two values based on validation performance. All other hyperparameters match those reported by~\cite{yang2024loretta}.}\label{a:tab:hparams:nlu-loretta}
\end{table}

\begin{table}[h!]
\setlength{\tabcolsep}{5pt}
\centering
\label{tab:model_hyperparameter_config}
\begin{tabular}{l | c}
\toprule
\textbf{Hyperparameter} & \textbf{Value} \\
\midrule
$\alpha$ & [0.5 1.0 2.0 8.0] \\
Learning Rate & [5e-4, 1e-3, 5e-3, 1e-2] \\
Scheduler & Linear \\
Optimizer & AdamW \\
Number of Epochs & 20 \\
Batch Size & 32 \\
Warmup Ratio & 0.06 \\
\bottomrule
\end{tabular}
\caption{Hyperparameter configurations for RoBERTa Base on the GLUE benchmark following~\cite{bershatsky2024lotr}. A grid-search  to set the learning rate and scale parameter for each task is conducted across the specified values.}\label{a:tab:hparams:nlu-lotr}
\end{table}

\begin{table*}[h!]
\setlength{\tabcolsep}{5pt}
\centering
\begin{tabular}{l | cccccc}
\toprule
Hyperparameter & SST-2 & MRPC & CoLA & QNLI & RTE & STS-B \\
\midrule
Optimizer & \multicolumn{6}{c}{AdamW} \\
Warmup Ratio & \multicolumn{6}{c}{0.06} \\
LR Schedule & \multicolumn{6}{c}{Linear} \\
Epochs & 10 & 40 & 40 & 20 & 40 & 20 \\
Learning Rate (Head) & 6E-3 & 3E-3 & 6E-3 & 2E-4 & 2E-3 & 2E-3 \\
Learning Rate (Encoder) & 1E-2 & 1E-2 & 1E-2 & 1E-2 & 2E-2 & 2E-2 \\
Batch Size & \multicolumn{6}{c}{32} \\
\bottomrule
\end{tabular}
\caption{Hyperparameter configurations for RoBERTa large on the GLUE benchmark. All other hyperparameters are taken from~\cite{2023vera}.}\label{a:tab:hparams:nlu-vera}
\end{table*}
\newpage
\subsection{Instruction tuning}\label{a:sec:hparams:inst}
For instruction tuning experiments we utilized Lightning AI's LitGPT codebase and training recipe. Hyperparameters are detailed below.
\begin{table}[h!]
\setlength{\tabcolsep}{5pt}
\centering
\label{tab:model_hyperparameter_config}
\begin{tabular}{l | c}
\toprule
\textbf{Parameter} & \textbf{Value} \\
\midrule
$\alpha$ & 16 \\
Learning Rate & 0.01 \\
Scheduler & Cosine \\
Optimizer & AdamW \\
Weight Decay & 0.01 \\
Number of Epochs & 1 \\
Steps & 51000 \\
Batch Size & 16 \\
Warmup Steps & 318 \\
\bottomrule
\end{tabular}
\caption{Hyperparameter configurations for LLama2-7B on the Alpaca dataset.}\label{a:tab:hparams:inst}
\end{table}

\subsection{DPO}\label{a:sec:hparams:dpo}
For preference optimization experiments we utilized using Huggingface trl library's dpo implementation and example script. Hyperparameters are detailed below.

\begin{table}[h!]
\setlength{\tabcolsep}{5pt}
\centering
\caption{Hyperparameter configurations for LLama2-7B on intel orca DPO pairs.}
\label{tab:new_model_hyperparameter_config}
\begin{tabular}{l | c}
\toprule
\textbf{Parameter} & \textbf{Value} \\
\midrule
$\alpha$ & 16 \\
Learning Rate & 0.00005 \\
Scheduler & Cosine \\
Optimizer & AdamW \\
Weight Decay & 0 \\
Number of Epochs & 1 \\
Batch Size & 16 \\
Warmup Steps & 200 \\
\bottomrule
\end{tabular}\label{a:tab:hparams:dpo}
\end{table}

\subsection{Protein Folding}\label{a:sec:hparams:folding}

For protein folding experiments, we utilized OpenFold \cite{Ahdritz2024OpenFold} training code and datasets. The following modifications were made to the ESMFold model architecture due to limited compute resources: a) utilize 12 Evoformer layers instead of the 48 used in \citep{Lin2021ESMFold} b) utilize ESM-2 35M instead of ESM-2 3B c) maintain outer product mean implementation from \citep{Jumper2021AlphaFold}. Optimizer and learning rate scheduler were identical to \citep{Jumper2021AlphaFold}. Models were trained for 850,000 steps with batch size of 32. Validation metrics were computed using the validation set from \citep{Ahdritz2024OpenFold}.

Preliminary experiments revealed that higher values of $\alpha$ yield better results in this setting. $\alpha$ for LoRA and LoRTA experiments was then selected in multiple stages. Initially, models were trained with $\alpha$ values of $256 \times r$ and $128 \times r$, and the best-performing model was chosen. If both configurations diverged, $\alpha$ was halved, and models were retrained with the next lower pair (e.g., $64 \times r$ and $32 \times r$). This halving process continued until a convergent model was found. See Table \ref{a:tab:hparams:folding} for the selected $\alpha$ values across experiments.

\begin{table}[h!]
\setlength{\tabcolsep}{5pt}
\centering
\caption{Selected $\alpha$ and LDDT-CA for protein folding models.}
\label{tab:folding}
\begin{tabular}{l | c | c}
\toprule
\textbf{Model} & \textbf{$\alpha$} & \textbf{Validation LDDT-C$\alpha$} \\
\midrule
LoRA (r = 1) & 128 & 0.668 \\
LoRTA (r = 64) & 128 & 0.663 \\
LoRTA (r = 8) & 256 & 0.667 \\
LoRTA (r = 1) & 2 & 0.656 \\
\bottomrule
\end{tabular}
\label{a:tab:hparams:folding}
\end{table}

\section{Additional results}~\label{a:sec:results}
\subsection{Instruction Tuning}

To further evaluate the fine-tuned models, we use MT-Bench~\citep{mtbench}, an LLM-as-a-judge benchmark. MT-Bench assesses multi-turn conversational and instruction-following abilities on 80 open-ended questions, covering diverse capabilities such as roleplaying, reasoning, coding and information retrieval. GPT-4 is used to score the outputs of the model on a scale of one to ten.

\begin{figure}[h!]
    \centering
    \begin{subfigure}[b]{0.45\textwidth}
        \centering
        \includegraphics[width=\textwidth]{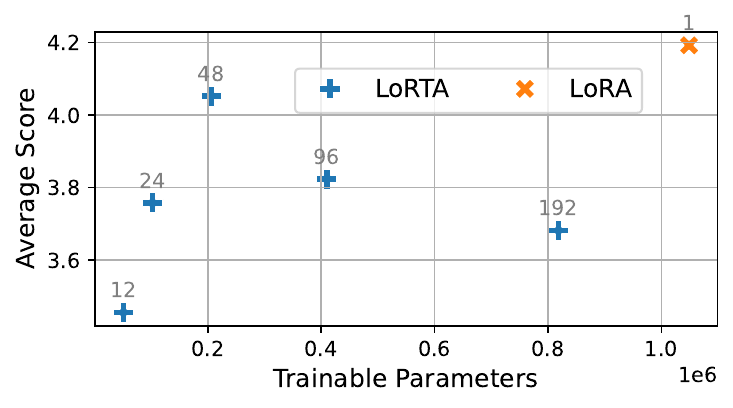}
    \end{subfigure}
    \hfill
    \begin{subfigure}[b]{0.45\textwidth}
        \centering
        \includegraphics[width=\textwidth]{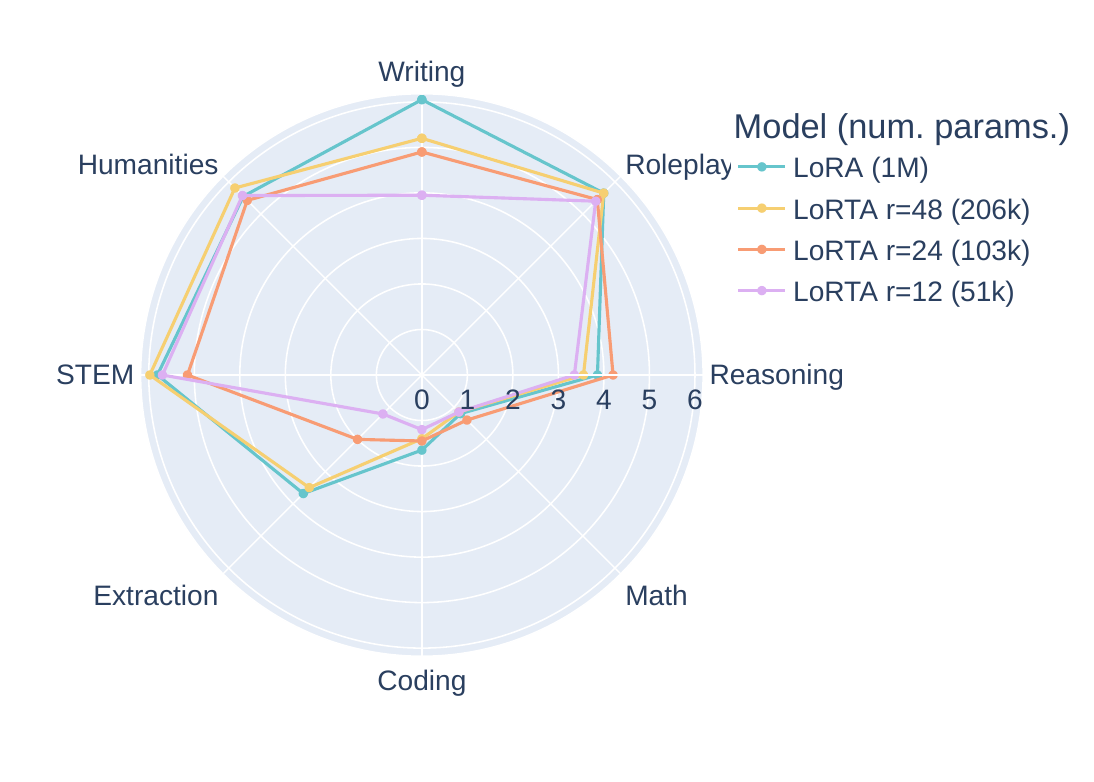}
    \end{subfigure}
    \caption{Performance on MT-Bench~\citep{mtbench} for Llama2-7b~\citep{touvron2023llama} models fine-tuned with LoRA and LoRTA. Higher is better. \textbf{Left}: Average score across all questions vs number of trainable parameters. Numbers on top of markers denote the adapter rank. \textbf{Right}: Average score per task.}
    \label{fig:mt-bench}
\end{figure}

As shown in Figure~\ref{fig:mt-bench}, LoRTA can almost match average performance despite using just 1/5th of the parameters (r=48). Unlike the loss observed in the Alpaca dataset, performance does not increase monotonically, potentially due to overfitting. Moreover, performance varies across tasks. For example, most LoRTA models surpass LoRA in reasoning but fall short in writing.
\subsection{Preference Optimization}
\begin{figure*}[h!]
\centering
\includegraphics[width=0.7\textwidth]{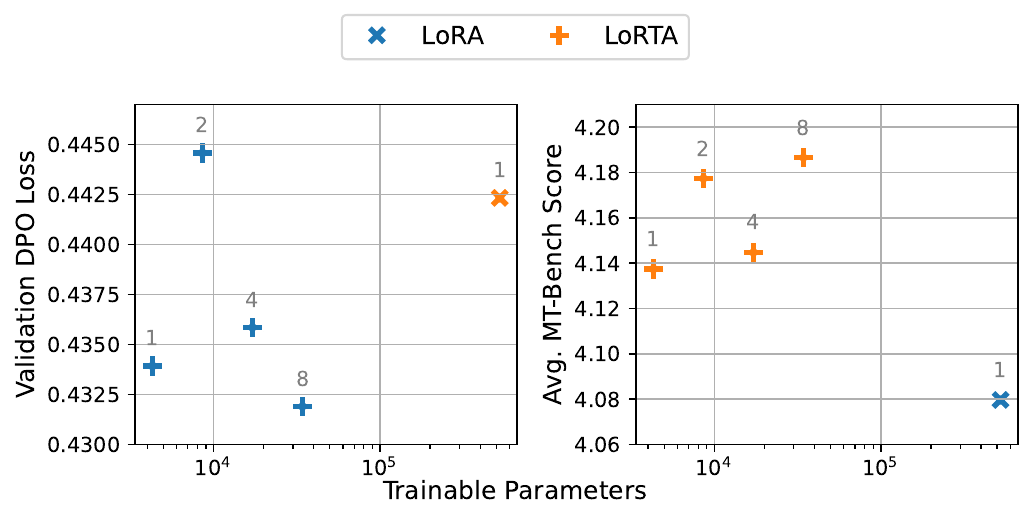}
    \caption{(Left) Mean DPO loss on held-out data from the orca dpo pairs dataset vs number of trainable parameters, lower is better. (Right) MT-Bench average scores Scores vs number of trainable parameters, higher is better.}
    \label{fig:dpo-loss-llama-7b}
\end{figure*}

As shown in Figure~\ref{fig:dpo-loss-llama-7b} LoRTA exhibited non-monotonic performance across ranks. This suggests that further hyperparameter tuning may be necessary to stabilize its performance. Although we did not tune hyperparameters, most ranks still outperformed LoRA with significantly fewer parameters. 

We further evaluated the fine-tuned models on the LLM-as-a-judge MT-benchmark. In this setting, LoRTA consistently outperformed LoRA across all ranks, including at rank 2 where it had shown higher DPO loss on the preference dataset. This improvement suggests enhanced out-of-distribution generalization capabilities for LoRTA adapters since MT-bench  differs from the training dataset.

\begin{figure}[h!]
    \centering
    \includegraphics[width=0.45\textwidth]{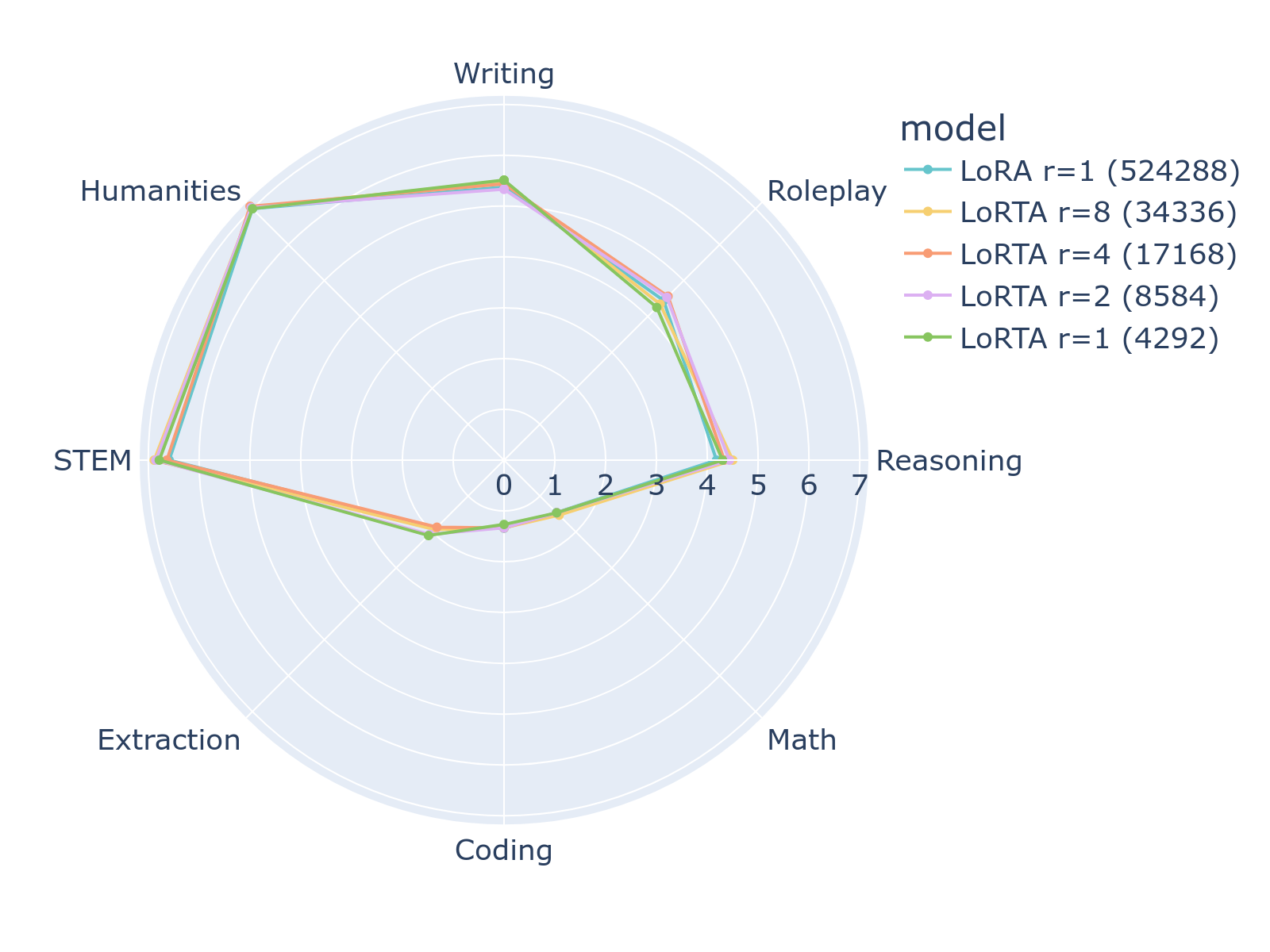}
    \caption{Performance on MT-Bench~\cite{mtbench} for llama2-7b~\cite{touvron2023llama} models fine-tuned with LoRA and LoRTA using dpo on intel orca pairs. Average score per task. Higher is better.}
    \label{fig:dpo-mt-bench-radar}
\end{figure}

\begin{figure*}[h!]
\centering
\includegraphics[width=0.7\textwidth]{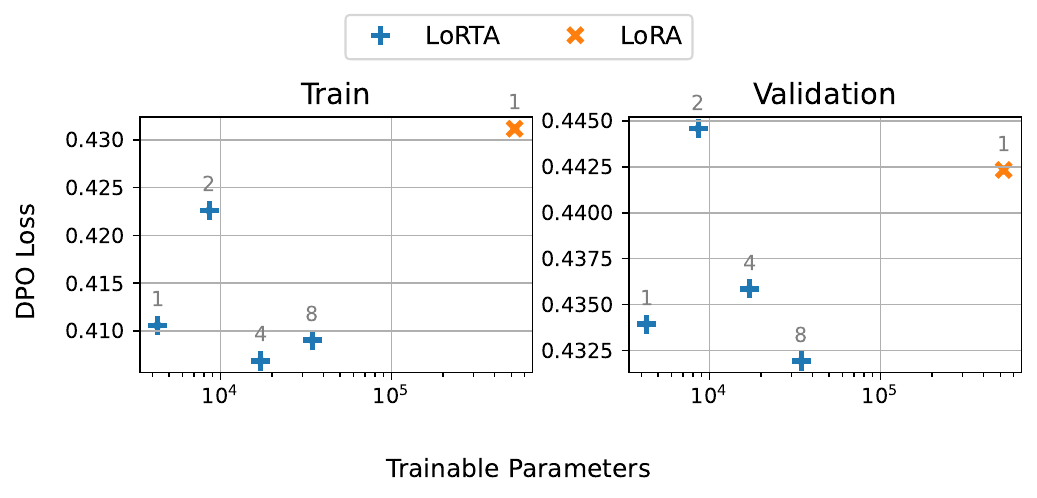}
    \caption{Mean DPO loss on the training (Left) and on held-out data (Right) from the orca dpo pairs dataset vs number of trainable parameters, lower is better.  }
    \label{fig:dpo-loss-llama-7b-train-val}
\end{figure*}

Figure~\ref{fig:dpo-loss-llama-7b-train-val} shows that Validation gains were primarily driven by reduced training error, though generalization slightly worsened, particularly at rank 2. On the other hand, as already mentioned, MT-bench performance was comparable o superior for LoRTA across all ranks, as shown in Figure~\ref{fig:dpo-mt-bench-radar}.

\subsection{Protein Folding}

\begin{figure*}[h!]
\centering
\includegraphics[width=0.7\textwidth]{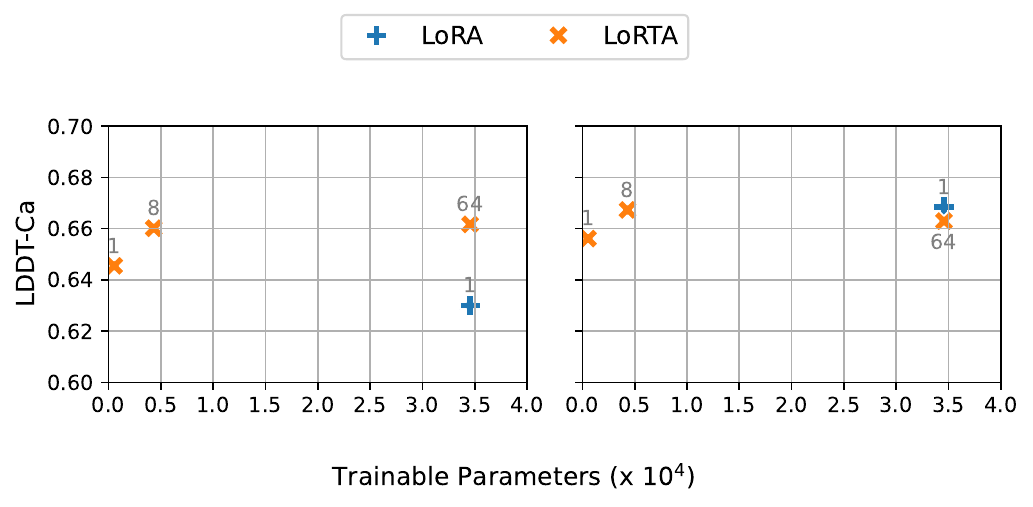}
    \caption{Mean LDDT-C$\alpha$ on held-out (left) and train sets. Higher is better. LoRTA rank 1 is competitive with LoRA rank 1 on the test set despite having 64x fewer parameters. Numbers on top of markers denote the adapter rank.}
    \label{fig:folding}
\end{figure*}

In figure~\ref{fig:folding} we include higher ranks for LoRTA in the protein folding experiment. However, note that increasing the rank beyond 1 and even matching the number of parameters in LoRA does not result in performance improvements. We also include train error to show that although LoRA rank 1 shows a performance improvement (Mean LDDT-C$\alpha$) in the training set, it shows a larger generalization gap.

\section{Training time and memory}~\label{a:sec:profiling}

During training, the reduction in GPU memory usage from shrinking optimizer states is marginal for parameter reductions beyond LoRA. Memory consumption in these cases is dominated by activations and caches stored during forward and backpropagation. Additional memory savings could be achieved by compressing activations or gradients, leveraging the low-rank structure of updates, or dynamically recomputing them. While our model features fewer trainable parameters and could theoretically benefit from the efficient tensor CP structure, such as faster training and lower memory usage, these advantages are not yet realized due to the limitations of our current implementation. We leave these optimizations for future work. However, the reduced parameter count already provides lower storage requirements and faster I/O.

 We conducted hardware profiling to compare the performance of our LoRTA implementation against LoRA using HuggingFace PEFT. The results demonstrate negligible differences in resource consumption between the two methods. The slight gap in training time for LoRTA can be addressed through further optimizations, ranging from leveraging tools like Torch Compile, to implementing our CP tensor adapter model more efficiently.

\begin{table*}[h!]
\setlength{\tabcolsep}{5pt}
\centering
\begin{tabular}{l | c | c | c | c | c}
\toprule
\textbf{Rank} & \textbf{Method} & \textbf{GPU Mem. (GB)} & \textbf{FLOPs (avg)} & \textbf{MACs (avg)} & \textbf{Time (s/step)} \\
\midrule
4 & LoRA & 12.84 & 272 & 136 & 0.07 \\
  & LoRTA & 12.88 & 272 & 136 & 0.14 \\
\midrule
64 & LoRA & 13.08 & 276 & 138 & 0.09 \\
   & LoRTA & 12.98 & 273 & 136 & 0.14 \\
\bottomrule
\end{tabular}
\caption{Maximum GPU memory usage (GB), average FLOPs(GB), MACs(GB), and training time (seconds per step) for LoRA and LoRTA.}
\label{tab:lorta_hardware}
\end{table*}

\end{document}